\documentclass[10pt,twocolumn,letterpaper]{article}

\usepackage{cvpr}              %

\usepackage{epsfig}
\usepackage{times,graphicx,tabulary,multirow,xspace}
\usepackage[dvipsnames]{xcolor}
\usepackage{amsmath,amssymb,soul,setspace,pifont}
\usepackage{makecell}
\usepackage[british,american]{babel}
\usepackage{url}
\usepackage[font=footnotesize,labelfont=bf]{caption}
\usepackage{appendix}
\usepackage{array,booktabs,calc}
\usepackage{placeins}
\usepackage[inline, shortlabels]{enumitem}
\usepackage{wrapfig}
\usepackage{paralist}
\usepackage{color}
\usepackage{colortbl}
\usepackage{inconsolata}
\usepackage{floatrow}
\usepackage{gensymb}
\usepackage{float}
\usepackage[accsupp]{axessibility}  %

\definecolor{citecolor}{RGB}{34,139,34}
\usepackage[pagebackref=true,breaklinks=true,letterpaper=true,colorlinks,citecolor=citecolor,bookmarks=false]{hyperref}

\usepackage[capitalize]{cleveref}
\crefname{section}{Sec.}{Secs.}
\Crefname{section}{Section}{Sections}
\Crefname{table}{Table}{Tables}
\crefname{table}{Tab.}{Tabs.}

\def\cvprPaperID{7746} %
\def\confName{CVPR}
\def\confYear{2022}
\newcommand{\shortname}{PlanarRecon\xspace}
\newcommand\blfootnote[1]{%
  \begingroup
  \renewcommand\thefootnote{}\footnote{#1}%
  \addtocounter{footnote}{-1}%
  \endgroup
}
\newcommand{\urlNewWindow}[1]{\href[pdfnewwindow=true]{#1}{\nolinkurl{#1}}}
\newcommand{\PAR}[1]{\vskip4pt \noindent{\bf #1~}}

\begin{document}

\title{\shortname: Real-time 3D Plane Detection and Reconstruction \\from Posed Monocular Videos}

\author{First Author\\
Institution1\\
Institution1 address\\
{\tt\small firstauthor@i1.org}
\and
Second Author\\
Institution2\\
First line of institution2 address\\
{\tt\small secondauthor@i2.org}
}

\author{
    Yiming Xie$^{1,2}$ 
    \quad Matheus Gadelha$^{3}$ 
    \quad Fengting Yang$^{4}$ 
    \quad Xiaowei Zhou$^{2\dagger}$
    \quad Huaizu Jiang$^{1\dagger}$ 
    \\
    $^1$Northeastern University \quad 
    $^2$Zhejiang University \quad 
    $^3$Adobe Research \quad
    $^4$The Pennsylvania State University \quad
}

\twocolumn[{%
\renewcommand\twocolumn[1][]{#1}%
\maketitle
\includegraphics[width=1\linewidth]{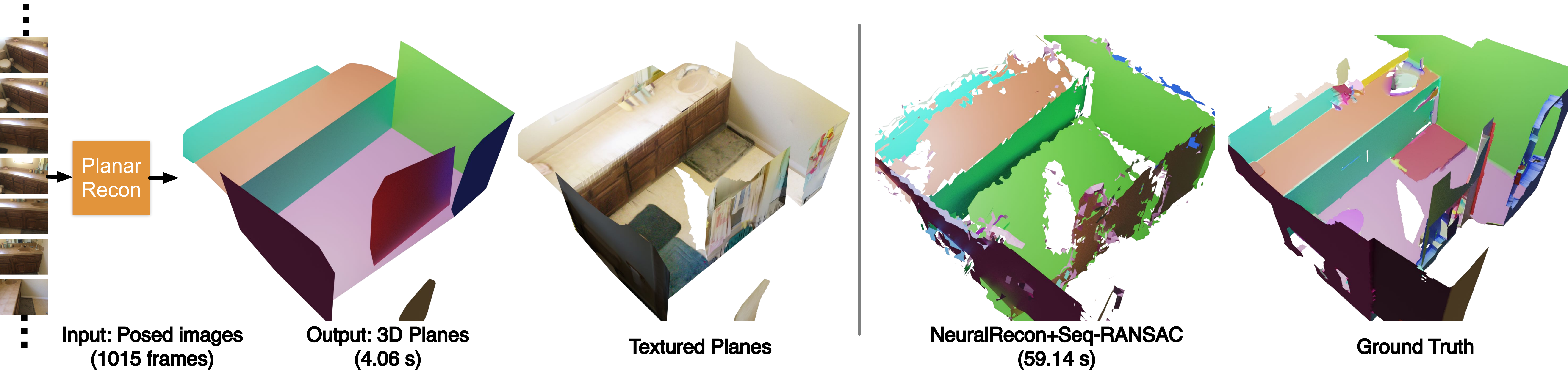}
    \captionof{figure}{
        \textbf{Comparison of the proposed approach, PlanarRecon, and our baseline.} 
        Colored planes mean different instances. 
        We also show the textured mesh for our approach.
        For the multi-view method NeuralRecon \cite{sun2021neuralrecon} + Seq-RANSAC \cite{fischler1981RANSAC}, the Sequential RANSAC is used to extract planes after the geometry reconstruction. Our model produces a much more accurate and coherent 3D plane detection in real-time. \emph{Best viewed in color.}
        \vspace{15pt}
    }
\label{fig:teaser}
}]

\maketitle

\begin{abstract}
   We present PlanarRecon -- a novel framework for globally coherent detection and reconstruction of 3D planes from a posed monocular video. 
Unlike previous works that detect planes in 2D from a single image, PlanarRecon incrementally detects planes in 3D for each video fragment, which consists of a set of key frames, from a volumetric representation of the scene using neural networks. 
A learning-based tracking and fusion module is designed to merge planes from previous fragments to form a coherent global plane reconstruction. 
Such design allows PlanarRecon to integrate observations from multiple views within each fragment and temporal information across different ones, 
resulting in an accurate and coherent reconstruction of the scene abstraction with low-polygonal geometry.
Experiments show that the proposed approach achieves state-of-the-art performances on the ScanNet dataset while being real-time. 
Code is available at the project page: \urlNewWindow{https://neu-vi.github.io/planarrecon/}.

\end{abstract}

\blfootnote{This work was partially done when Yiming Xie was a research assistant at Zhejiang University. $^\dagger$ Corresponding authors: Xiaowei Zhou and Huaizu Jiang.}

\section{Introduction}
\label{sec:intro}

Recovering 3D planar surfaces from a posed monocular video is a critical task for many downstream applications in 3D vision, such as Augmented and Virtual Reality (AR and VR), interior modeling, and human-robot interaction. 
Planar surfaces provide a compact representation and important geometric cues of the 3D scene. 
AR, for example, an accurate, consistent, and real-time 3D plane detection is required to enable realistic and immersive interactions between AR effects and the surrounding physical scene.
While state-of-the-art visual-inertial SLAM systems~\cite{camposORBSLAM3AccurateOpenSource2020, qinGeneralOptimizationbasedFramework2019a, AugmentedRealityApple} can accurately track camera poses, real-time image-based 3D plane detection remains a challenging problem due to low detection quality and high computation demands.

Most recent deep learning-based plane recovery works \cite{liuPlaneNetPieceWisePlanar2018,liuPlaneRCNN3DPlane2018, yangRecovering3DPlanes2018, yuSingleImagePiecewisePlanar2019, tanPlaneTRStructureGuidedTransformers2021}  focus on the single-view case. Their pipeline typically aims to jointly segment the plane instances and regress the plane parameters (\ie, surface normals and offsets). Despite the significant progress made in this direction using deep neural networks, because of the single-view scale ambiguity, these methods cannot deliver absolute depth estimation in the unknown scenes. Moreover, fusing the plane detections from multiple views is not trivial. Jin \etal\cite{jin2021planar} extend the single-view approach~\cite{liuPlaneRCNN3DPlane2018} to sparse views (mostly 2 views), where time-consuming energy minimization is used to fuse single-view detections.
As their design does not consider the temporal consistency, however, it is still unclear how to extend this work to video inputs,  which are more natural and common vision sources for applications like AR and VR.

In the traditional 3D vision, a few attempts~\cite{Bartoli07, BaillardZ99, ManhattanworldMVS,SinhaSS09,ConchaC15, YangSKS16, YangS19 } have been made to recover planes from multi-view images and videos. But they usually rely on hand-crafted features~\cite{BaillardZ99, Bartoli07, SinhaSS09} and/or strong geometric priors of the scene~\cite{ManhattanworldMVS,ConchaC15, YangSKS16, YangS19}. In real-world scenarios, these features may be unreliable and such priors may not always hold due to the scene complexity, such as lighting condition change, textureless regions, fixtures violating the Manhattan world assumption~\cite{coughlan2000manhattan}, etc. An efficient plane recovery method that is robust to the aforementioned challenges and makes no strong assumptions of the scene is desired.

\begin{figure*}[ht]
    \centering
       \includegraphics[width=\linewidth]{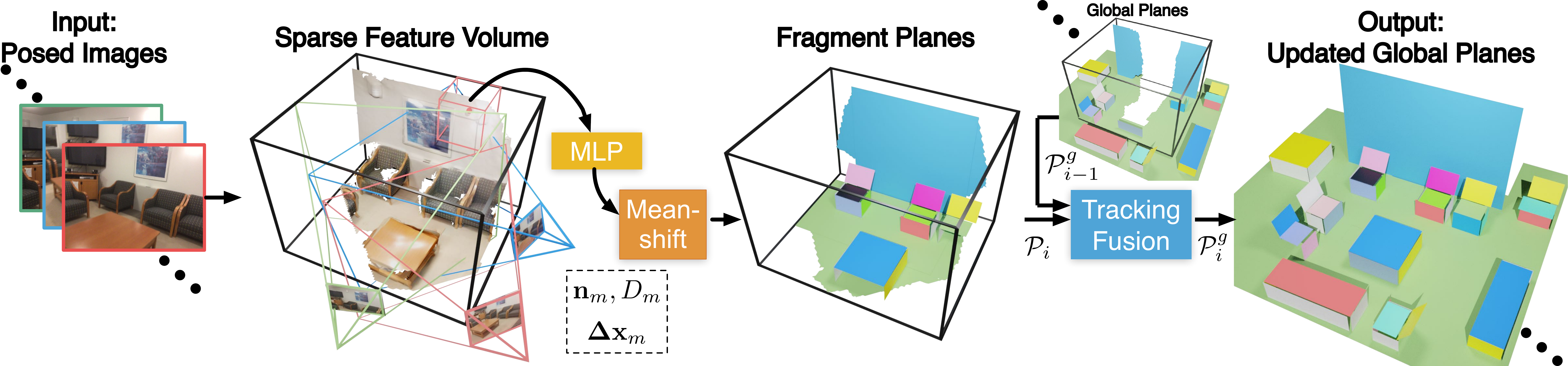}
       \caption{
           \textbf{\shortname architecture.}
           \shortname backprojects the image features into a fragment bound volume $\mathcal{F}_i$ and gradually sparsify the volume in a coarse-to-fine approach to form a sparse feature volume. 
           A MLP network will be used to predict plane parameters $[\mathbf{n}_m,D_m]$ and votes $\Delta \mathbf{x}_m$ for each voxel. 
           Then these hybrid geometric primitives $[\mathbf{n}_m, \mathbf{x}_m' = \mathbf{x}_m + \Delta \mathbf{x}_m]$ are passed through a Mean-shift clustering to form plane instance $\mathcal{P}_i$ in fragment bound volume $\mathcal{F}_i$. 
           The tracking and fusion module will match 3D planes $\mathcal{P}_i$ in current fragment bounding volume and global planes $\mathcal{P}_{i-1}^g$ from previous fragments. 
           The matched plane pairs will be refined, yielding the final 3D plane reconstruction.
           }
       \label{fig:arch}
\end{figure*}

To meet this demand, we propose a novel learning-based framework, called \textbf{PlanarRecon}, to perform 3D plane detection and reconstruction in real-time from a posed monocular video. The main idea of our proposed PlanarRecon is illustrated in Fig. \ref{fig:arch}. 
It consists of two major components. 
The first component is  \emph{fragment-based plane detection}. Given a video input, we sequentially split it into multiple non-overlapping fragments.
For each fragment, PlanarRecon constructs 3D feature volumes by back-projection of the image features, which fuses information from multiple views. 
Based on the occupancy classification, for each occupied voxel, we estimate the plane parameters as well as its displacement to shift it to the geometric centroid of the plane it belongs to. Mean-shift clustering~\cite{yuSingleImagePiecewisePlanar2019} is then performed to group voxels that have similar plane parameters and shifted positions to get plane detections in the fragment. The second component is  \emph{plane tracking and fusion}. PlaneRecon maintains a global reconstruction of planes using plane detections from all previous fragments. When a new fragment is processed, we resort to the attention mechanism~\cite{vaswani2017attention} to compute the similarities between the global reconstructed planes and the current detections. A differentiable Hungarian matching algorithm is then 
used to obtain the correspondences of planes. The global reconstruction is updated accordingly to ensure temporal coherence.

Our model incrementally obtains 3D plane reconstructions from the input video. 
Thanks to its fast inference speed, the system can run in real-time, enabling more authentic interaction experiences with the scene for a downstream AR application, for instance. 
Compared with single-image based approaches~\cite{liuPlaneNetPieceWisePlanar2018,liuPlaneRCNN3DPlane2018, yangRecovering3DPlanes2018, yuSingleImagePiecewisePlanar2019, tanPlaneTRStructureGuidedTransformers2021}, PlanarRecon directly regresses planes from 3D feature volumes. 
It not only fuses information from multiple views, but also offers a coherent reconstruction of the scene without the scale ambiguity. 
On the other hand, compared with traditional multi-view reconstruction approaches, our learning-based model is more robust to scene complexity~\cite{BaillardZ99, Bartoli07, SinhaSS09} and does not rely on the existence of certain scene priors~\cite{ManhattanworldMVS,ConchaC15, YangSKS16, YangS19}. 
Experimental results on the ScanNet benchmark~\cite{dai2017scannet} show that PlanarRecon achieves state-of-the-art accuracy.

To summarize, our main contributions are twofold: 
\textit{i}) We propose PlanarRecon to detect and reconstruct 3D planes from a posed monocular video. 
To our best knowledge, PlanarRecon is the first learning-based approach in this direction. 
\textit{ii}) We propose a novel volume-based plane reconstruction approach that can detect, track, and fuse plane instances, directly in 3D. %
Our model integrates observations from multiple frames and temporal information from the video, leading to globally coherent plane detection and reconstruction. 
Compared with existing approaches, our approach is more robust and runs significantly faster.

\section{Related Work}
\label{sec:related}
\paragraph{Multi-view Plane Reconstruction.}
There is a long line of research on multi-view plane reconstruction from sequences of frames with known camera poses. 
Early works usually first perform sparse 3D reconstruction with point~\cite{Bartoli07} or line features~\cite{BaillardZ99} and then group the sparse 3D representations with certain heuristics. 
However, the reconstruction accuracy is heavily dependent on the hand-crafted features,  and not robust to other factors like lighting changes and textureless regions. 
Other works pose this problem as an image segmentation task.
There are methods that assign each pixel to one of the plane hypotheses in an MRF formulation~\cite{ManhattanworldMVS, SinhaSS09}.
Others extend this framework to handle non-planar surfaces~\cite{GallupFP10} 
and introduce superpixel segmentation to better tackle the textureless regions~\cite{ConchaC15}. 
Our approach, unlike the existing ones, employs convolutional neural networks (CNNs) to extract features and perform optimization in a data-driven manner. 
To the best of our knowledge, we are the first to perform multi-view plane reconstruction using a deep learning approach.

Another line of work takes monocular videos as input to simultaneously estimate the camera poses and reconstruct the planar surfaces in a SLAM fashion. 
However, these works commonly assume that the world consists of a horizontal ground and a few vertical planes (\eg, facades or walls)~\cite{YangSKS16, YangS19}, and/or the planar structures only exist in low-gradient areas~\cite{ConchaC15}. 
In our work, we do not make such assumptions, leading to a more applicable approach.

\paragraph{Learning-based Plane Reconstruction.}
While we are not aware of any existing deep learning-based work that reconstructs piece-wise planes from multi-view images or video sequences, there are a few studies aiming to recover the planar structures from one or two views. Several works treat the plane reconstruction from single views as an instance segmentation problem~\cite{liuPlaneNetPieceWisePlanar2018,liuPlaneRCNN3DPlane2018, yangRecovering3DPlanes2018, yuSingleImagePiecewisePlanar2019, tanPlaneTRStructureGuidedTransformers2021}. 
Deep networks are employed to jointly predict the plane instance segmentation and plane parameters. The segmentation masks are later projected into 3D using the predicted parameters for plane reconstruction. To further improve the reconstruction accuracy, other works detect and enforce the inter-plane relationships among their plane instances~\cite{qian2020learning} or perform segmentation on horizontal and vertical planes separately with panorama inputs~\cite{sun2021indoor}.

The problem becomes more challenging in the two-view case. 
To generate a unified 3D reconstruction, the model has to correctly associate the plane instances across frames.
Previous work has proposed learning descriptors for each instance and matching them with certain optimization algorithms~\cite{shi2018planematch, jin2021planar}. 
However, it is still unclear how well this approach can be generalized to multi-view cases. Instead, we directly perform plane detection, tracking, and fusion in 3D, which reduces the ambiguity of the matching process and leads to higher reconstruction accuracy and more coherent results.

\paragraph{Learning-based Multi-view Stereo.}
Our work is also related to recent deep learning-based multi-view stereo (MVS) methods~\cite{yao2018mvsnet,yao2019recurrent,cheng2020deep, yang2020cost,ma2021epp}. 
One representative work in this direction is MVSNet~\cite{yao2018mvsnet}. For each frame, MVSNet selects a few neighboring frames as sources and uses a plane-sweeping mechanism to stack per-frame features into a 4D cost volume which is later aggregated with a 3D CNN in order to predict the depth map of the reference frame. 
The final 3D model is reconstructed by fusing all the estimated depth maps. 
There are a few other works that do not rely on the plane-sweeping mechanism.
Some approaches are inspired by the patch match algorithm~\cite{bleyer2011patchmatch}, which randomly initialize a large number of depth proposals and propagate the ones with low photometric errors calculated with deep features~\cite{wang2021patchmatchnet, lee2021patchmatch}.
Sinha \etal first generate sparse depth maps with SuperPoint~\cite{detone2018superpoint} and later perform depth completion~\cite{sinha2020deltas}. 

All the works use depth maps as the intermediate 3D representation and only consider the information from neighboring frames.
Several methods~\cite{houMultiViewStereoTemporal2019,liu2019neural,duzceker2021deepvideomvs} integrate temporary information from a video sequence.
Other works take the truncated signed distance function (TSDF) as 3D representation and project per-frame features into a pre-defined volumetric space for TSDF regression~\cite{murez2020atlas, sun2021neuralrecon}.
In recent work~\cite{wei2021nerfingmvs}, all frames' information is incorporated with neural radiance fields~\cite{mildenhall2020nerf} for multi-view depth estimation. 
Our work adopts the framework from~\cite{sun2021neuralrecon} for the initial 3D geometry estimation due to its high efficiency. 
But different from it, our method directly produces 3D plane reconstruction, which results in less complicated geometry and thus more friendly for downstream tasks that require high processing throughput (\eg, AR, physical simulation, interactive applications in general).
\newcommand{\calF}{\mathcal{F}}
\newcommand{\vn}{\mathbf{n}}
\newcommand{\vx}{\mathbf{x}}
\newcommand{\vf}{\mathbf{f}}
\newcommand{\vh}{\mathbf{h}}
\newcommand{\vM}{\mathbf{M}}
\newcommand{\vS}{\mathbf{S}}

\section{Method}
\label{sec:method}
Given a sequence of monocular video frames and their camera poses, we first find a set of suitable key frames sequentially from the incoming image stream. 
Following \cite{houMultiViewStereoTemporal2019}, a new incoming frame is selected as a key frame if its relative translation is greater than $t_{max}$ or the relative rotation angle is greater than $R_{max}$ when compared with the last selected key frame.
When the number of key frames reaches $N_k$, the sequence of $N_k$ consecutive key frames forms a local fragment. 
Each fragment is defined as $\calF_i=\{I_{i,j} \}_{j=1}^{N_k}$, where $i$ is the fragment index, $j$ is the key frame index.
Given these sequential fragments, our goal is to incrementally detect and reconstruct 3D planes that approximate the underlying 3D scene geometry. 
Here, we define a plane as a planar structure from a single object instance in the same way as proposed by Liu \etal~\cite{liuPlaneRCNN3DPlane2018}. 

Fig.~\ref{fig:arch} presents an overview of the proposed method. 
We divide our pipeline into two major components. In Section~\ref{subsec:Frag}, we introduce fragment-based plane detection, where we first obtain plane detections for each fragment. Section~\ref{subsec:Fusion} shows a plane tracking and fusion module, where we integrate plane detections from the current fragment and previous fragments together to form a coherent 3D plane reconstruction of the scene in an incremental manner.

\begin{figure}[t]
    \centering
       \includegraphics[width=1.0\linewidth]{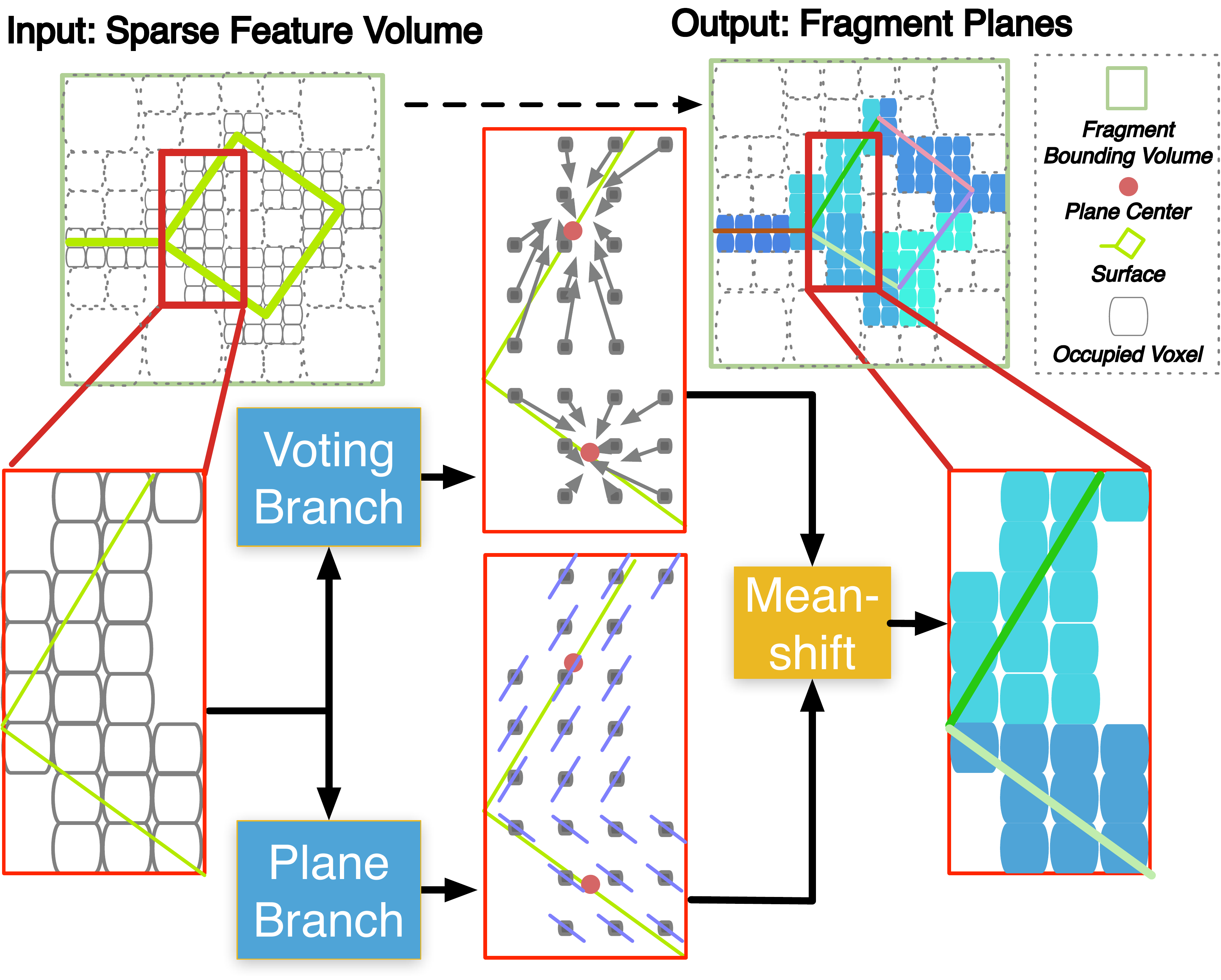}
       \caption{
            \textbf{2D illustration of fragment-based plane detection.}
            Note that we ignore the sparse volume reconstruction, which is detailed in the supplementary material. 
            The colored grids in the fragment bound volume mean different plane instances they belong to. 
            \emph{Best viewed in color.}
           }
       \label{fig:detect}
\end{figure}

\subsection{Fragment-based Plane Detection}
\label{subsec:Frag}

\PAR{Overview.} The task of extracting planes would be significantly easier if accurate geometry were
available.
While detailed geometric information is hard to obtain, removing the majority of the empty space is relatively simple. 
Thus, we adopt a coarse-to-fine approach to build 3D sparse feature volumes, where we classify each voxel as occupied or not (\ie, if it belongs to the surface or not). For each occupied voxel, we add two sibling branches -- a plane branch and a voting branch.
Their goal is to estimate the plane parameters and displacement from the centroid of the plane it belongs to, respectively. 
Finally, clustering is performed on the occupied voxels to group the ones that have similar plane parameters and displacements together in order to get plane detections in 3D. 
An illustration of this procedure is shown in Fig.~\ref{fig:detect}.

\PAR{3D Sparse Feature Volume Construction.} We define the 3D space for $\calF_i$ as the region within a cubic-shaped bounding volume that encloses the view-frustums of all the key frames $\{I_{i,j} \}$. 
We obtain per-voxel features for occupancy classification by passing each of the key frames $I_{i,j}$ through a 2D CNN backbone. 
A 3D feature volume is obtained by back-projecting those image-based features into the fragment bounding volume $\calF_i$.
Average pooling is performed to aggregate features for the same voxel from multiple pixels across different views.
Next, we use 3D sparse convolutions to efficiently process the feature volumes and estimate voxel-wise occupancy scores. 
We define that a voxel is occupied if it is within  $\lambda$ distance from the surface.
A voxel whose occupancy score is lower than a threshold $\theta$ is considered as void space and therefore is removed from the following plane detection process.
Similar to previous volumetric 3D reconstruction works~\cite{sun2021neuralrecon,ji2020surfacenet_plus}, after this sparsification process, the fragment bounding volume $\calF_i$ is upsampled twice in the next level, and this process
is repeated in a coarse-to-fine manner. In our implementation, we use three levels of sparse volumes. We provide more details in the supplementary material.

\begin{figure}[t!]
    \centering
       \includegraphics[width=0.9\linewidth]{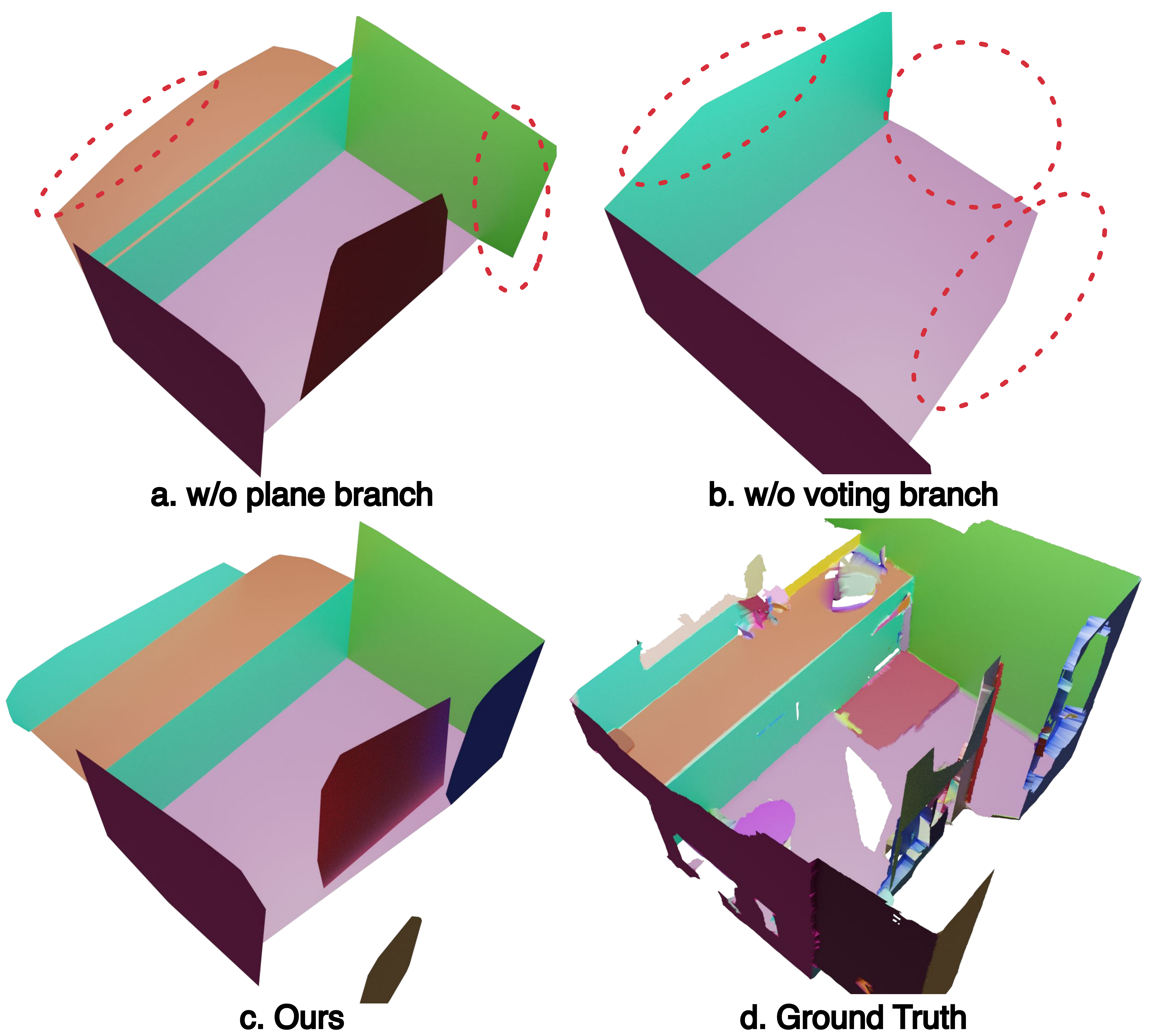}
       \caption{
           \textbf{Effectiveness of the plane and voting branches.} Colored planes mean different instances. 
           The results in (a), (b) fail to separate planes and produce inaccurate reconstruction (as shown in the \textbf{\color{red}red} circle).
           \emph{Best viewed in color.} 
           }
       \label{fig:ablation}
\end{figure}

\PAR{Plane Branch.} After the occupancy analysis, the model would obtain a set of features $\{[\vx_m, \vf_m]\}_{m=1}^M$ for the $M$ occupied voxels\footnote{For simplicity, we omit the fragment index here.},
where $\vx_m \in \mathbb{R}^3$ is the center position of a voxel. $\vf_m \in \mathbb{R}^C$ is a $C$-dimensional feature obtained from the sparse volume constructed in the previous stage, which will be used to regress plane parameters, including the surface normal $\vn_m\in \mathbb{R}^3$ and plane offset $d_m\in \mathbb{R}^1$, associated to the voxel. 

For the surface normal, similar to~\cite{liuPlaneRCNN3DPlane2018}, we maintain a set of \emph{anchor normals}. For each occupied voxel, we first predict the anchor normal that is closest to the ground-truth value and then regress a 3D residual vector from it. We found it works better than directly regressing the normal values. As our model directly works in 3D, unlike~\cite{liuPlaneRCNN3DPlane2018}, which obtains anchor normals by clustering the ground-truths in 2D, our anchor normals can directly be defined in the world coordinate system. In specific, we manually design six anchor normals, where two of them are parallel to the ground plane and the rest anchor normals is perpendicular to the ground.

For the plane offset regression, we estimate the distance $D_m \in \mathbb{R}^1$ from the center position of a voxel $x_m$ to the plane it belongs to. Given an estimated plane normal \textbf{$n_m$} and distance $D_m$, the plane offset $d_m$ can be obtained via
\begin{align}
    &d_m = -\langle \tilde{\vx}_m, \vn_m\rangle,~~\tilde{\vx}_m = \vx_m + D_m \vn_m,
\end{align}
where $\langle\cdot,\cdot\rangle$  denotes the inner product between two vectors.

The plane branch is instantiated by multi-layer perceptions (MLPs). 
We use the cross-entropy loss to supervise the learning of anchor normal selection, and the \texttt{smooth\_L1} loss to supervise the anchor normal residual vector and plane offset distance regression.

\PAR{Voting Branch.}
Given the plane parameter estimations, we can already group voxels together to get plane detections. However, such a strategy may fail to separate two planar surfaces that have similar plane parameters even they are far from each other, \eg two tables surfaces that have the same height. To this end, inspired by~\cite{qi2019deep}, we also estimate the displacement $\Delta \vx_m \in \mathbb{R}^3$ from the voxel center $\vx_m$ to the centroid of the planar surface it belongs to. With the estimated displacement $\Delta \vx_m$, we can shift the voxel $\vx_m$ to the new position $\vx_m' = \vx_m + \Delta \vx_m$.  
The predicted 3D displacement $\Delta \vx_m'$ is supervised by L1 loss.

\PAR{Voxel Clustering.}
Once we have geometric primitives for every single voxel (shifted voxels $\vx_m' \in \mathbb{R}^3$ and surface normals $\vn_m\in \mathbb{R}^3$), we group the voxels to form plane instances in the local fragment volume $\calF_i$. 
We found that the plane offset $d_m$ is not useful for the voxel clustering and introduces an extra computation burden. 
Following \cite{yuSingleImagePiecewisePlanar2019}, we use an efficient mean-shift clustering algorithm. We use the centers of the clusters as the final plane parameters for each plane instance. 
We show clustering results with and without using the plane and voting branches in Fig.~\ref{fig:ablation}.

\begin{figure}[t]
    \centering
       \includegraphics[width=1.0\linewidth]{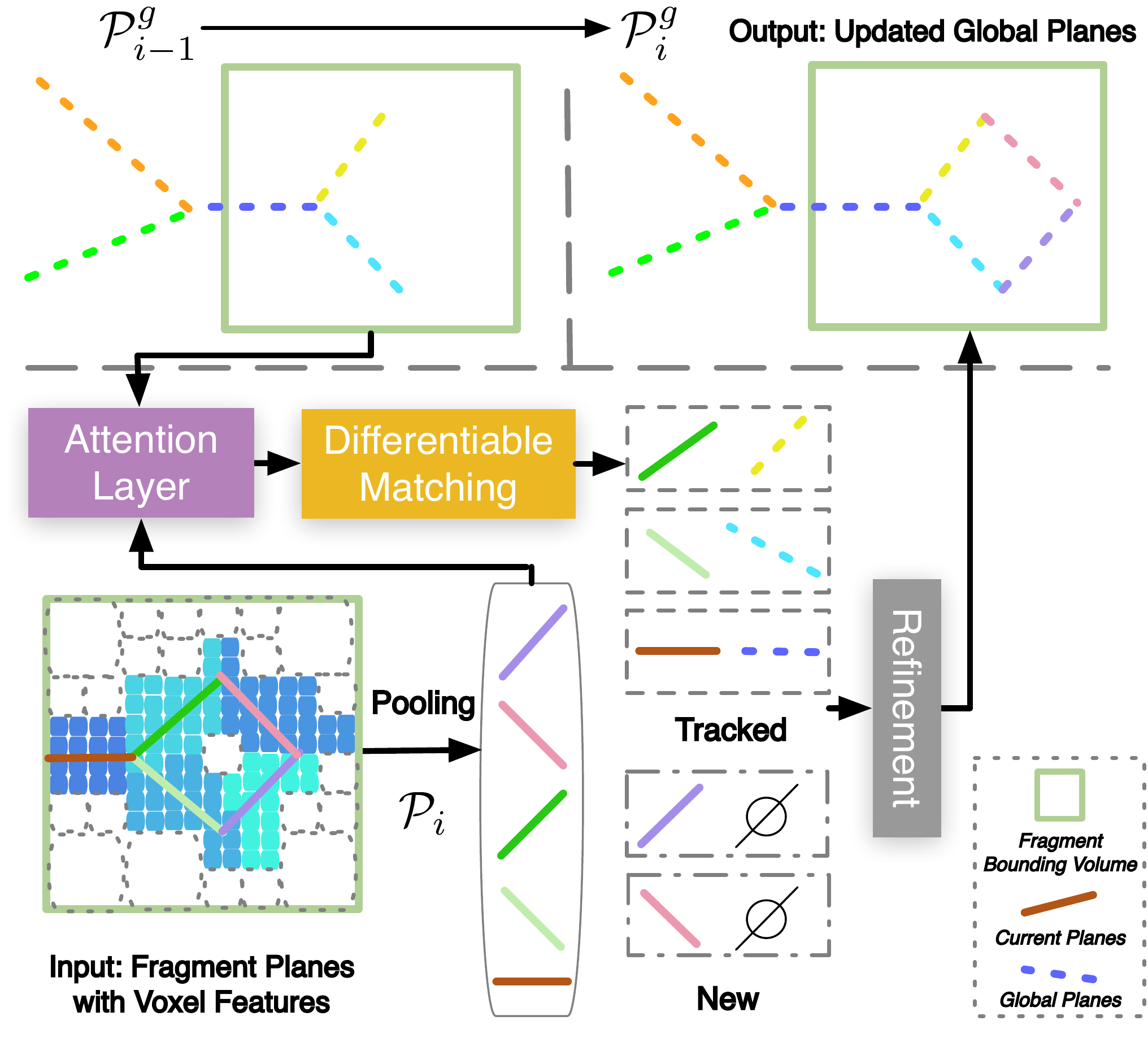}
       \caption{
            \textbf{2D illustration of 3D plane tracking and fusion.}
            The colored line segments mean different plane instances with their descriptors. 
            \emph{Best viewed in color.}
           }
       \label{fig:track_fuse}
\end{figure}
\newcommand{\calP}{\mathcal{P}}
\subsection{Plane Tracking and Fusion}
\label{subsec:Fusion}
\noindent\textbf{Overview.} To integrate plane detections from different fragments to form a globally coherent 3D plane reconstruction, we design a learning-based tracking and fusion module. We keep a global plane reconstruction $\calP_i^g$ consisting of a set of plane instances $\calP_i^g=\{P_{i,m}^g\}$, where $\calP_0^g=\emptyset$ and $m$ is the plane instance index. 
As shown in Fig. \ref{fig:track_fuse}, given the plane detections $\calP_i=\{P_{i,n}\}$ from the fragment $\calF_i$, we need to integrate $\calP_{i-1}^g$ and $\calP_i$ to get an updated global plane reconstruction $\calP_i^g$. It involves two steps. First of all, we do \emph{plane tracking} -- we need to find matchings between planes of $\calP_{i-1}^g$ and $\calP_i$. Then, we \emph{fuse} matched planes to get refined plane reconstructions. 
Each of the two steps is introduced in detail next.

\PAR{Differentiable Matching for Plane Tracking.} To find correspondences of two sets of plane instances, $\calP_{i-1}^g$ and $\calP_i$, we first need to compute their similarity scores $\vS$.
Inspired by \cite{sarlin20superglue}, we found that integrating the contextual cues from all plane instances may boost matching accuracy. We therefore resort to the self-attention mechanism~\cite{vaswani2017attention} to allow message passing between different plane instances. For the plane instance $P_{i,n}$, in addition to using its plane parameters $\vn_{i,n}$ and $d_{i,n}$ (obtained from the cluster center), we also use the average pooled volume features $\bar{\vf}_{i,n}$ from all the voxels in the cluster as an input to the attention module. We use self-attention and cross-attention graph networks, similar to~\cite{sarlin20superglue}, to get augmented feature vectors for the plane instances $P_{i-1,m}^g$ and $P_{i,n}$ as $\vh_{i-1,m}^g$ and $\vh_{i,n}$, respectively. The similarity score between them is then defined as
\begin{align}
    \vS(m, n) = \langle \vh_{i-1,m}^g, \vh_{i,n} \rangle.
\end{align}
For plane tracking, we need to find an optimal matching matrix $\vM^*$, where $\vM(m, n)\in\{0, 1\}$ such that
\begin{align}
\vM^* = \arg\min_{\vM} \sum_{m,n}\vS(m,n)\vM(m,n).     
\end{align}
The optimal matchings $\vM^*$ can be efficiently solved using the Sinkhorn algorithm~\cite{marco2013sinkhorn,Sinkhorn1967ConcerningNM}, which is differentiable. We introduce its loss function in the supplementary material. 
As shown in Fig.~\ref{fig:track_fuse}, in practice, it is possible that some plane instances in $\calP_{i-1}^g$ may not find their correspondences because a particular plane instance is not visible in the new fragment. 
We keep it in the updated global reconstruction $\calP_{i}^g$. On the other hand, if a plane in $\calP_i$ does not find its correspondence, it is likely that this is a new plane instance that is observed for the first time. Thus we initialize a new plane in $\calP_{i}^g$.

\PAR{3D Plane Refinement for Plane Fusion.}
To better leverage the temporal information, we refine plane parameters for the matched plane instances.
Suppose $P_{i-1, m}^g$ and $P_{i,n}$ are two matched plane instances, the plane instance is updated as 
\begin{align}
    & P_{i,m}^g = \frac{\gamma P_{i-1, m}^g + P_{i,n}}{\gamma + 1},
\end{align}
where $\gamma$ is a parameter controlling the updating speed provided by a GRU. 
The GRU fuses plane features $\bar{\vf}_{i,n}$ with $\bar{\vf}_{i-1,m}^g$ and produces the updated plane features $\bar{\vf}_{i,m}^g$, which will be passed through the MLP layers to predict $\gamma$.
We slightly abuse notations here, $P_{i,m}^g$ can be either surface normal $\vn_{i,m}^g$, plane offset $d_{i,m}^g$, or pooled volume feature vector $\bar{\vf}_{i,m}^g$.

\begin{table*}[t]
    \centering
    \caption{\textbf{3D geometry metrics on ScanNet.} 
    Our method outperforms the compared approaches by a significant margin in almost
    all metrics.
    $\uparrow$ indicates bigger values are better, $\downarrow$ the opposite.
    The best numbers are in bold.
    We use two different validation sets following Atlas \cite{murez2020atlas} (top block) and  PlaneAE \cite{yuSingleImagePiecewisePlanar2019} (bottom block).
    }
    \label{tab:scannet-3d}
    \resizebox{1.0\textwidth}{!}{
    \begin{tabular}{cccccc>{\columncolor[gray]{0.902}}ccc}
    \Xhline{3\arrayrulewidth}
    Method             & validation set                                     & Comp $\downarrow$ & Acc $\downarrow$           & Recall $\uparrow$         & Prec $\uparrow$           & \textbf{F-score} $\uparrow$  & Max Mem. (GB) $\downarrow$    & Time ($ms$/keyframe) $\downarrow$   \\ \hline
    NeuralRecon \cite{sun2021neuralrecon} + Seq-RANSAC & \multirow{4}{*}{Atlas~\cite{murez2020atlas}}    & 0.144           & 0.128           & 0.296           & 0.306           & 0.296    & \textbf{4.39}       & 586   \\ 
    Atlas \cite{murez2020atlas} + Seq-RANSAC  &              & \textbf{0.102}           & 0.190  & 0.316           & 0.348           & 0.331  & 25.91 & 848     \\ 
    ESTDepth \cite{Long_2021_CVPR} + PEAC \cite{feng2014fast}& &  0.174  & 0.135         & 0.289  & 0.335           & 0.304     & 5.44     & 101   \\ 
    Ours &                                                    & 0.154           & \textbf{0.105}   & \textbf{0.355}           & \textbf{0.398}  & \textbf{0.372} & 4.43 & \textbf{40}  \\

    \Xhline{3\arrayrulewidth}
    PlaneAE \cite{yuSingleImagePiecewisePlanar2019} & \multirow{2}{*}{PlaneAE~\cite{yuSingleImagePiecewisePlanar2019}}           & \textbf{0.128}           & 0.151           & 0.330           & 0.262           & 0.290   & 6.29        & \textbf{32}     \\ 
    Ours          &                                         & 0.143  & \textbf{0.098}  & \textbf{0.372}  & \textbf{0.412}  & \textbf{0.389} & \textbf{4.43} & 40     \\ 
    \Xhline{3\arrayrulewidth}
    \end{tabular}
    }
\end{table*}
\section{Experiments}
\label{sec:exp}
In this section, we conduct a series of experiments to evaluate the 3D plane detection and reconstruction quality as well as different design considerations of PlanarRecon.

\subsection{Setup}
\PAR{Datasets.} 
We perform the experiments on ScanNetv2~\cite{dai2017scannet}. 
The ScanNetv2 dataset contains RGB-D videos taken by a mobile device from 1613 indoor scenes. The camera pose is associated with each frame.
As no ground truths are provided in test set, we follow PlaneRCNN \cite{liuPlaneRCNN3DPlane2018} and generate 3D plane labels on the training and validation set. 
Our method is evaluated on two different validation sets with different scene splits used in previous works~\cite{murez2020atlas,liuPlaneNetPieceWisePlanar2018}.  

\PAR{Evaluation Metrics.}
We evaluate the performance of our method in terms of the 3D plane detection, which can be evaluated using instance segmentation metrics following previous works, and 3D reconstruction. For plane instance segmentation, due to the geometry difference between the ground truth and prediction meshes, we follow the semantic evaluation method proposed in~\cite{murez2020atlas}. 
More specifically, given a vertex in the ground truth mesh, we first locate its nearest neighbor in the predicted mesh and then transfer its prediction label. 
We employed three commonly used single-view plane segmentation metrics~\cite{yangRecovering3DPlanes2018,liuPlaneRCNN3DPlane2018, tanPlaneTRStructureGuidedTransformers2021,yuSingleImagePiecewisePlanar2019} for our evaluation: rand index (RI), variation of information (VOI), and segmentation covering (SC).
We also evaluate the geometry difference between predicted planes and ground truth planes. 
We densely sample points on the predicted planes and evaluate the 3D reconstruction quality using 3D geometry metrics presented by Murez \etal~\cite{murez2020atlas}.

\PAR{Baselines.}
Since there are no previous work that focus on learning-based multi-view 3D plane detection, we compare our method with three types of approaches:
(1) single-view plane recovering~\cite{yuSingleImagePiecewisePlanar2019}; 
(2) multi-view depth estimation~\cite{Long_2021_CVPR} + depth-based plane detection~\cite{feng2014fast}; and
(3) volume-based 3D reconstruction \cite{sun2021neuralrecon, murez2020atlas} + Sequential RANSAC~\cite{fischler1981RANSAC}.

Since baselines (1) and (2) predict planes for each view, we add a simple tracking module to merge planes predicted by the baseline in order to provide a fair comparison. 
The tracking and merging process we designed for our baselines is detailed in the supplementary material.
We use the same key frames as in PlanarRecon for baselines (1) and (2).
For (3), we first employ~\cite{sun2021neuralrecon, murez2020atlas} to estimate the 3D mesh of the scene, and perform sequential RANSAC to group the oriented vertices of the mesh into planes. 
Please refer to our supplementary material for the details of the sequential RANSAC algorithm.
For \cite{sun2021neuralrecon, murez2020atlas}, we run sequential RANSAC every time when a new 3D reconstruction is completed to achieve incremental 3D plane detection. 
\begin{table}[t]
    \centering
    \caption{\textbf{3D plane segmentation metrics on ScanNet.} 
    Our method also outperforms competing approaches in almost all metrics when evaluating
    plane segmentation metrics.
    $\uparrow$ indicates bigger values are better, $\downarrow$ the opposite.
    The best numbers are in bold.
    We use two different validation sets following Atlas \cite{murez2020atlas} (top block) and  PlaneAE \cite{yuSingleImagePiecewisePlanar2019} (bottom block).
    }
    \label{tab:segmt_eval}
    \resizebox{1.0\textwidth}{!}{
        \begin{tabular}{ccccc}
            \Xhline{3\arrayrulewidth}
            Method                                              & VOI $\downarrow$                    & RI $\uparrow$      & SC $\uparrow$    \\
            \hline
            NeuralRecon \cite{sun2021neuralrecon} + Seq-RANSAC  & 8.087                               & 0.828                & 0.066               \\
            Atlas \cite{murez2020atlas} + Seq-RANSAC            & 8.485                               & 0.838                & 0.057      \\
            ESTDepth \cite{Long_2021_CVPR} + PEAC \cite{feng2014fast} & 4.470                          & 0.877                & 0.163                 \\ 
            Ours                                                & \textbf{3.622}                      & \textbf{0.897}       & \textbf{0.248}               \\
            \Xhline{3\arrayrulewidth}
            PlaneAE \cite{yuSingleImagePiecewisePlanar2019}        & 4.103                               & \textbf{0.908}                 & 0.188              \\
            Ours                                                & \textbf{3.622}                      & 0.898      & \textbf{0.247}     \\
            \Xhline{3\arrayrulewidth}
        \end{tabular} }
\end{table}

\subsection{Results}
\begin{figure*}[ht]
    \vspace{-1.2cm}
    \centering
       \includegraphics[width=1\linewidth]{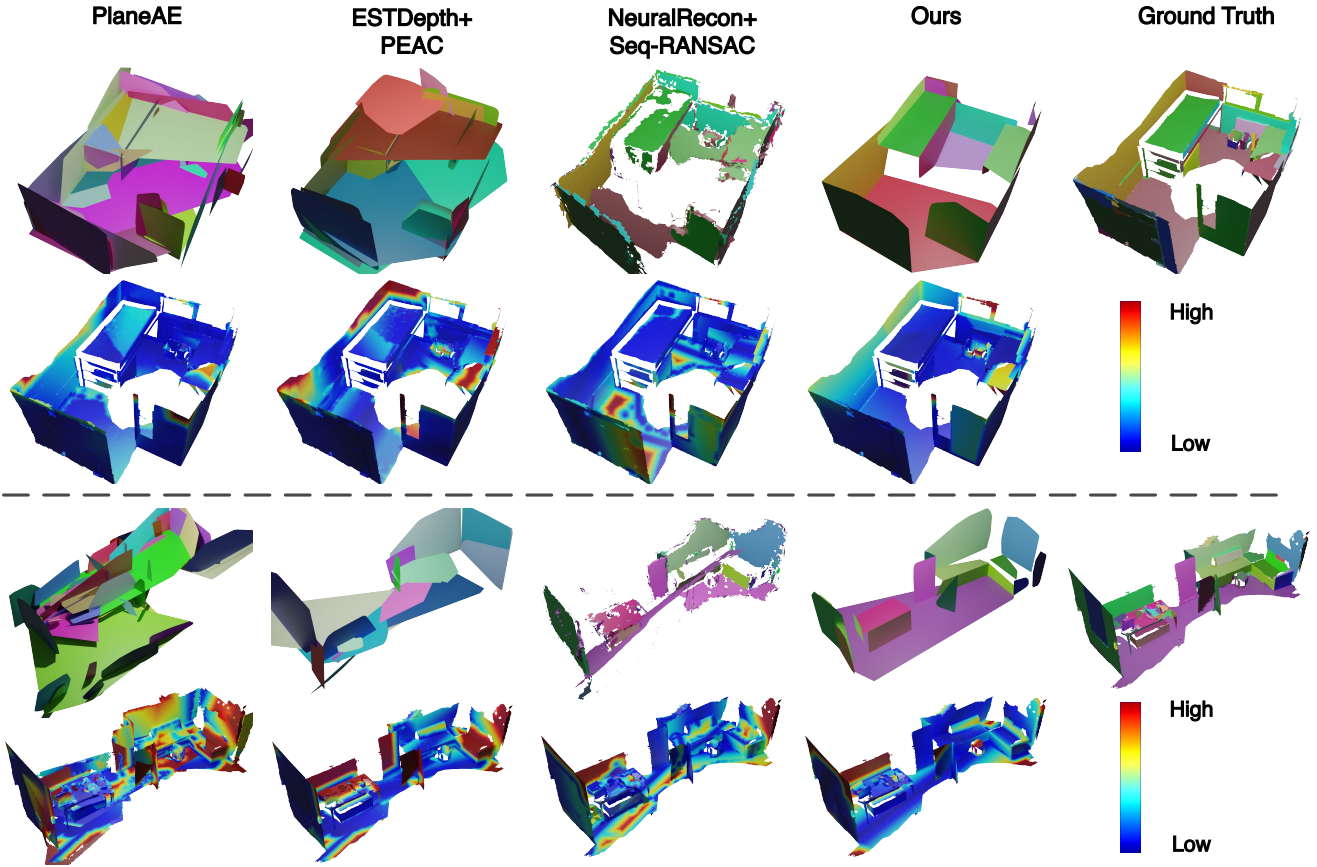}
       \caption{
           \textbf{Qualitative results on ScanNet.}
        Colored planes mean different instances. 
        We present the error map for each method. %
        When compared to other methods, \shortname is capable of producing consistent sets of
        planes that better summarize and accurately represent the scene geometry.
        \emph{Zoom in for details}.}
       \label{fig:qualitative}
\end{figure*}

\PAR{Geometric Accuracy.} 
The 3D geometry evaluation results are shown in Tab.~\ref{tab:scannet-3d}. 
Our method produces better performance than both single-view methods and multi-view methods.
We believe the improvements come from volume-based 3D plane detection followed by a learning-based tracking and fusion module. 
Compared to single-view methods, PlanarRecon produces more accurate and globally coherent 3D planes as can be seen in Fig.~\ref{fig:qualitative}. 
Our method has much higher performance than its single-view counterparts in terms of accuracy (Acc.), recall, precision (Prec.), and F-score. 
In comparison with the depth-based multi-view method, we outperform the baseline on all the geometry metrics. 
The volume representation allows our method to capture the local smoothness priors, leading to locally coherent results compared to estimating geometry per frame and then fuse them later in the depth-based method. 
Besides, the plane tracking module, which can be jointly learned with the 3D plane detector, is more robust than a hand-crafted tracking mechanism, resulting in globally coherent 3D plane detection results. 
Our method also surpasses the volumetric baselines in terms of accuracy, recall, precision, and F-score. 
The improvements potentially come from the fact that the model can be end-to-end trained, leading the network to learn features that better adapt to scene complexity.

\PAR{Instance Segmentation Accuracy.} The 3D plane instance segmentation evaluation results are presented in Tab.~\ref{tab:segmt_eval}. 
Our method outperforms the baselines with sequential RANSAC across all the metrics.  Sequential RANSAC tends to mis-group planar surfaces that are parallel and close to each other.
For the single-view method, since it processes each frame independently, more plane instances are actually detected, which leads to a higher recall and so higher RI.
However, its performance is worse than ours in terms of VOI and SC. Because these two metrics focus on information variation and segmentation overlapping between prediction and ground truth, which focus more on the overall segmentation quality.

\PAR{Efficiency.}
We also report the average running time of the baselines and our method in Tab.~\ref{tab:scannet-3d}. 
Only the inference time on key frames is computed. 
Like in \cite{sun2021neuralrecon}, for volumetric methods (Atlas, NeuralRecon, and ours), the running time is obtained by dividing the time of the number of key frames in the local fragment. 
The running time for PlanarRecon is measured on an NVIDIA V100 GPU. 
When comparing to NeuralRecon + sequential RANSAC, Atlas + sequential RANSAC and ESTDepth + PEAC, we use the running time reported in their paper~\cite{sun2021neuralrecon, murez2020atlas, Long_2021_CVPR, feng2014fast, yuSingleImagePiecewisePlanar2019}.
As shown in Tab.~\ref{tab:scannet-3d}, our runtime is 40ms per key frame, which corresponds to a frame rate of 25 key frames per second and outperforms most of our baselines. 
Specifically, our method runs $\sim$\textbf{2$\times$ faster} than ESTDepth \cite{Long_2021_CVPR} + PEAC \cite{feng2014fast} and $\sim$\textbf{15$\times$ faster} than NeuralRecon \cite{sun2021neuralrecon} + Seq-RANSAC. 
Similar to~\cite{sun2021neuralrecon}, we use volumetric representations, which can remove redundant computation in depth-based multi-view or single-view methods.
Compared to other volume-based methods, our method can directly predict planes via a neural network, avoiding time-consuming RANSAC.

We report the maximum GPU memory usage in the inference stage in Tab.~\ref{tab:scannet-3d}. 
Because our method uses the sparse volume as representation of scene, 
the GPU memory cost is a function of the surface area of the scene. Based on our experiments on ScanNet, PlanarRecon can reconstruct scenes using up to 4.43 GB GPU memory. 
In the training stage, PlanarRecon uses up to 22.87 GB GPU memory.

\begin{table}[t]
    \centering
    \caption{\textbf{Ablation studies.} 
    We report 3D geometry metrics F-score and 3D plane segmentation metrics on ScanNet.
    }
    \label{tab:ablation}
    \resizebox{1.0\textwidth}{!}{
        \begin{tabular}{ccccc}
            \Xhline{3\arrayrulewidth}
            Method                                              &F-score $\uparrow$ & VOI $\downarrow$                    & RI $\uparrow$      & SC $\uparrow$    \\
            \hline
            Ours, full approach                            & \textbf{0.372} & \textbf{3.622}                      & \textbf{0.897}       & \textbf{0.248}               \\
            \rowcolor[gray]{0.902}w/o plane branch          & 0.349    & 3.839                                   & 0.883         & 0.226                   \\
            \rowcolor[gray]{0.902}w/o voting branch          & 0.148    & 4.970                                   & 0.715         & 0.154                   \\
            \rowcolor[gray]{0.902}w/o learning-base fusion & 0.358 & 3.798                                & 0.897                & 0.222                   \\
            \rowcolor[gray]{0.902}w/o learning-base tracking  & 0.362 & 3.639                             & 0.893                & 0.247                   \\
            \Xhline{3\arrayrulewidth}
        \end{tabular} }
\end{table}

\PAR{Qualitative Results.}
We provide the qualitative results in Fig.~\ref{fig:qualitative}. 
To visualize NeuralRecon, we project the vertices to the planes they belong to and keep the edge of the mesh.
For other methods, we can first get points from the depth map (baselines) or voxel center (ours).  
Then we project the point to the planes they belong to and generate the mesh using Delaunay triangulation.

Notice how our method produce a smaller amount of planes that better summarize
the geometry of the scene. 
NeuralRecon + sequential RANSAC generates coherent results because the sequential RANSAC runs on the entire reconstructed mesh in each time step, which is \emph{significantly} more time-consuming.
Besides, this RANSAC-based approach is sensitive to the hyper-parameters choice. 
The ground in the third row is split into two planes because of the relatively small distance threshold. %
But if we set it with a larger value, some close planes tend to merge into one instance.
PlaneAE and ESTDepth + PEAC suffer from obtaining coherent results. Due to the inconsistent predictions among different views, it is hard to match their predictions along with the whole video, leading to many false plane detections and inaccurate scene geometry. 

\subsection{Ablation Study}
We also conduct several ablation experiments on the ScanNet dataset. 
The ablation study is shown in Tab.~\ref{tab:ablation}.

\PAR{Plane branch.}
To validate the plane branch, we only use the shifted voxels $\mathbf{x}_m'$ to group voxels to form plane instances in local fragment volume $\mathcal{F}_i$. 
As shown in Tab.~\ref{tab:ablation}, the full approach can separate different plane instances accurately. 
Comparing visualization results (a) and (c) in Fig.~\ref{fig:ablation} also shows that the plane reconstruction result without the plane branch is prone to producing inaccurate planes.

\PAR{Voting branch.} 
We validate the voting branch by removing this module. 
As shown in Tab.~\ref{tab:ablation} and Fig.~\ref{fig:ablation}(b), both geometry accuracy and segmentation accuracy drop rapidly. 
The results demonstrate that the voting branch is effective to get accurate 3D plane segments, especially when two planar surfaces are in the same plane. 

\PAR{Learning-based Tracking and Fusion.}
We track planes directly using the IoU of two planes instead of learning-based tracking. 
The IoU measures the intersection over a union of two sets of voxels which correspond to two planes. 
As shown in Tab.~\ref{tab:ablation}, the F-score has 1\% improvement when the learning-based tracking is used. 
The 3D instance segmentation results are improved due to more robust matches.
We also removed the learning-based fusion and instead directly use a fixed weight $\gamma$.
As shown in Tab.~\ref{tab:ablation}, the 3D geometry metrics are improved with the fusion module.

\subsection{Limitations and Failure Cases}
Our approach assumes the existence of planar structures in the scene, which are very common in indoor scenes. 
But in scenarios where such planar surfaces do not exist, for instance, nature scenes like mountains, forests, etc., our approach may fail. 
We aim to build a planar simplification of the scene, which unavoidably introduces certain deviations if the surfaces are not perfectly planes. 
As shown in Fig.~\ref{fig:qualitative}, the error mainly comes from the small planar areas and the areas on the edges. %

\section{Conclusion}
\label{sec:conclusion}
In this work, we introduced a novel system, PlanarRecon, for real-time 3D plane detection and reconstruction with posed monocular video. 
The key idea is to use a volumetric representation of the scene and learning-based tracking and fusion module to detect, match and fuse planes in 3D for each video fragment incrementally. 
This design enables PlanarRecon to produce accurate and globally coherent 3D planes in real-time. 
Experiments show that PlanarRecon outperforms state-of-the-art methods and runs in real-time. 
The global 3D planes detected by PlanarRecon can be directly used in downstream applications like AR/VR. 

\PAR{Acknowledgement.}
YX and HJ would like to acknowledge the support from Google Cloud Program Credits.

{\small
\bibliographystyle{ieee_fullname}
\bibliography{egbib.bib}
}

\end{document}